%% file: main.tex
\documentclass[accepted]{uai2023} %

\usepackage[american]{babel}

\usepackage{natbib} %
\usepackage{mathtools} %
\usepackage{booktabs} %
\usepackage{tikz} %

\usepackage{amsmath}
\usepackage{amsthm}
\usepackage{amssymb}
\usepackage{bm}
\usepackage{subcaption}
\usepackage[normalem]{ulem}
\usepackage[makeroom]{cancel}
\usepackage{amsfonts}
\usepackage{xcolor}
\usepackage{graphicx}
\usepackage{tikz}
\usepackage{booktabs}
\usepackage{array}
\usepackage{multirow}
\usepackage{enumitem}
\usepackage{algorithm}
\usepackage{algorithmic}
\usepackage{setspace}
\usepackage{xfrac}
\usepackage{xr}

\renewcommand{\paragraph}[1]{\textbf{#1}~~}

\usetikzlibrary{shapes.geometric}
\usetikzlibrary{arrows.meta}
\usetikzlibrary{bayesnet}

\definecolor{mydarkblue}{rgb}{0,0.08,0.45}
\usepackage[capitalize,nameinlink]{cleveref}
\crefname{section}{Sec.}{Secs.}
\crefname{appendix}{App.}{Apps.}
\crefname{algorithm}{Alg.}{Algs.}
\creflabelformat{equation}{#2\textup{#1}#3}

\definecolor{darkgreen}{rgb}{0.0, 0.5, 0.0}

\input{macros.tex} %

\title{Exploiting Inferential Structure in Neural Processes}

\author[1]{\href{mailto:<d.v.tailor@uva.nl>?Subject=Exploiting Inferential Structure in Neural Processes}{Dharmesh Tailor}{}}
\author[2]{Mohammad Emtiyaz Khan}
\author[1]{Eric Nalisnick}
\affil[1]{%
    University of Amsterdam\\
    Amsterdam\\
    Netherlands
}
\affil[2]{%
    RIKEN Center for AI Project\\
    Tokyo\\
    Japan
}

\begin{document}
\maketitle

\begin{abstract}
    Neural Processes (NPs) are appealing due to their ability to perform fast adaptation based on a context set.  This set is encoded by a latent variable, which is often assumed to follow a simple distribution. However, in real-word settings, the context set may be drawn from richer distributions having multiple modes, heavy tails, \etc  In this work, we provide a framework that allows NPs' latent variable to be given a rich prior defined by a graphical model.  These distributional assumptions directly translate into an appropriate aggregation strategy for the context set.  Moreover, we describe a message-passing procedure that still allows for end-to-end optimization with stochastic gradients. 
    We demonstrate the generality of our framework by using mixture and Student-\emph{t} assumptions that yield improvements in function modelling and test-time robustness.
\end{abstract}

\section{Introduction}\label{sec:intro}

\begin{figure*}[!t]
    \centering
    \begin{subfigure}[b]{.45\linewidth}
        \centering
        \includegraphics{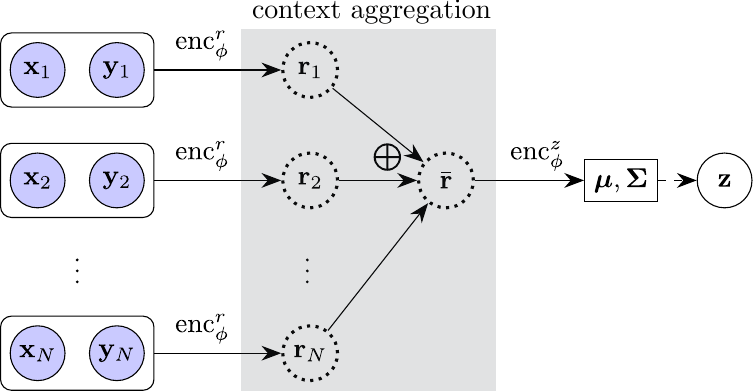}
        \caption{Sum-Decomposition Network}
        \label{fig:sdn}
    \end{subfigure}
    \begin{subfigure}[b]{.45\linewidth}
        \centering
        \includegraphics{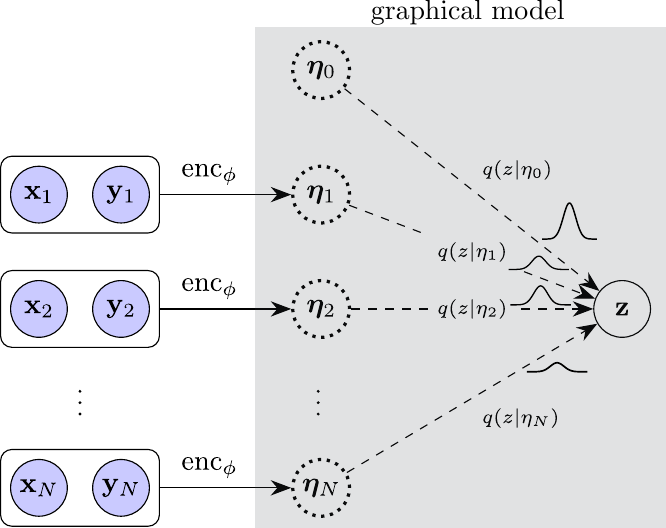}
        \caption{Structured Inference Network}
        \label{fig:sin}
    \end{subfigure}
    \caption{A set of context points $(\vx_i,\vy_i)$ can be aggregated using a sum-decomposition architecture, shown on the left. A shared network $\text{enc}_{\phi}^r$ is used for all
    followed by a pre-specified aggregation operation that pools all embeddings. The output is passed to a further network $\text{enc}_{\phi}^z$. We propose an alternative, shown on the right, that does not require an additional network and rather uses inference on a probabilistic graphical model (PGM) to automatically aggregate the embeddings. Different types of exponential-family distributions and mixtures, denoted here by $q(z|\eta_i)$, can be used resulting in different aggregation mechanisms.}
    \label{fig:archit}
\end{figure*}

Many real-world tasks require models to make predictions in new scenarios on short notice.  For example, climate models are often asked to make predictions at novel locations \citep{vaughan2022convolutional}. Data is collected in well-populated regions but predictions for remote regions (\eg~mountain ranges, forests \etc) are desirable as well.  Neural processes (NPs) \cite{garnelo2018neural} are models designed for situations such as this.  At test time, the model is seeded with a context data set from the target setting that (hopefully) allows the NP to make accurate predictions despite the possibly novel conditions.  This behavior is implemented using an efficient encoder architecture that scales linearly with respect to the size of the context set and is permutation-invariant to its order (see \cref{fig:sdn}).  Unfortunately, NPs still have shortcomings that make them brittle for this ambitious use case.  For example, NPs commonly underfit \citep{kim2019attentive} and suffer from limited representation power \citep{wagstaff2019limitations}.  Previous work has attempted to fix these problems by enriching the encoder's architecture, \eg attention \citep{kim2019attentive}, transformers \citep{nguyen2022transformer}, and convolutions \citep{gordon2019convolutional, foong2020meta}.

We consider an alternative approach that incorporates the structure and assumptions of the data into NPs' latent variable.  To accomplish this, we consider priors defined by a probabilistic graphical model (PGM).  While incorporating rich PGMs may seem like it would hinder the scalability that makes NPs an attractive model, we show it does not.  Using \textit{structured inference networks} \citep{lin2018variational} (see \cref{fig:sin}), we can still train NPs end-to-end, using variational message passing for the PGM \citep{winn2005variational} and stochastic gradients for the neural networks.  Within our framework, encoding the context set becomes analogous to inference in the PGM.  This means that the aggregation operation over the context set is completely and automatically determined by the choice of PGM prior.  We show that---under a simple Gaussian PGM---our approach recovers \textit{Bayesian Aggregation} (BA) \citep{volpp2020bayesian}. %

In this paper, we describe a general framework for placing PGM priors on NPs with latent variables.  We primarily focus on cases that exhibit conditional conjugacy but provide some discussion of the fully non-conjugate case as well.  We show the explicit updates for mixture priors and Student-\emph{t} assumptions.  In the experiments, we show NPs with these mixture and heavy-tail assumptions---which we term \emph{mixture} and \emph{robust} Bayesian Aggregation, respectively---demonstrate improved performance in regression and image completion.

\section{Background}\label{sec:bg}
\paragraph{Problem setup} NPs assume a partition of the data into a \textit{context} set and a \textit{target} set. The former is used by the model to seed adaptation.  The latter is a set of points from the same domain as the context set and for which we will make predictions.  Specifically, we denote the context set for the $l$\textsuperscript{th} task as $\data_c^{(l)} = \{\vx_{c,i}^{(l)}, \vy_{c,i}^{(l)} \}_{i=1}^{N_{l,c}}$, where $\vx$ denotes a feature vector and $\vy$ the corresponding response vector.  The target set for the $l$\textsuperscript{th} task is denoted similarly as $\data_{t}^{(l)} = \{\vx_{t,i}^{(l)}, \vy_{t,i}^{(l)} \}_{i=1}^{N_{l,t}}$.  At test time, for a new task $l_{*}$, we observe $\data_{c}^{(l_{*})}$ and $\{\vx_{t,i}^{(l_{*})}\}_{i=1}^{N_{l_{*},t}}$.  The target responses $\{\vy_{t,i}^{(l_{*})}\}_{i=1}^{N_{l_{*},t}}$ are unobserved, and our goal is to predict them.

\begin{figure}[t]
\centering
\resizebox{0.6\linewidth}{!}{
\input{figs/tikz/generative.tex}
}
\caption{Generative process for multi-task learning with context/target splits along with the variational approximation (\protect\tikz[baseline=-0.5ex,inner sep=0pt]{\protect\draw[->, dashed, >={stealth'},line width=1pt,shorten <=0.75pt,shorten >=0.75pt, red] (0,0) -- ++(0.55,0);}).}
\label{fig:mtl_generative}
\end{figure}
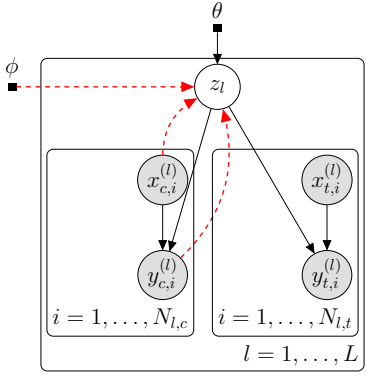

\paragraph{Neural Processes} 
\textit{Neural processes} \citep{garnelo2018neural} frame few-shot learning as a multi-task learning problem \citep{heskes2000empirical},
employing a conditional latent variable model with context/target splits on task-specific datasets as shown in \cref{fig:mtl_generative}.
Training amounts to maximising the following \emph{conditional} marginal likelihood across $L$ tasks:
\begin{align}
    &\sum_{l=1}^L \log p_{\theta} (\data_{t}^{(l)} \mid \data_{c}^{(l)}) \nonumber \nonumber \\ 
    &\gtrapprox \sum_{l=1}^L \E_{q_{\phi}(\vz | \data_c^{(l)} \cup \data_t^{(l)})} \sqr{ \log p_{\theta} (\data_{t}^{(l)} \mid \vz ) } \label{eq:elbo} \\
    &\textstyle\quad - \KL{q_{\phi}(\vz | \data_c^{(l)} \cup \data_t^{(l)})}{q_{\phi}(\vz | \data_c^{(l)})} \nonumber
\end{align}
where $\vz_l$ is the task-specific latent variable and $\theta$ is the global parameter that is shared across tasks.
The marginalization over task-specific latent variables is typically intractable hence approximate inference is used,
\begin{align}
    p_{\theta}(\vz \mid \data_c) &= \textstyle\frac{1}{p_{\theta}(\data_c)} \prod_{i=1}^{N_c} \nonumber
    p_{\theta}(\vy_{c,i} \mid \vx_{c,i}, \vz) \; p_{\theta}(\vz)  \label{eq:post_true} \\
    &\approx q_{\phi}(\vz \mid \data_c).
\end{align}
where we have dropped task indices for notational simplicity.
The variational approximation is amortised, meaning a recognition network is used.
For a Gaussian approximation, the mean and variance are parameterized by neural networks (NNs) that take as input sets of datapoints: $q_{\phi}(\vz \mid \data_c) = \N(\vz \mid \widetilde{\vmu}, \widetilde{\MSigma})$ 
with $\big(\widetilde{\vmu},\widetilde{\MSigma}\big) = \textrm{enc}_{\phi}(\data_c)$.
Throughout we assume covariance matrices have diagonal structure, resulting in fully-factorized Gaussian distributions.

\paragraph{Neural Process Extensions}
NPs come in two variants, the aforementioned (latent) NP \citep{garnelo2018neural} and the conditional NP (CNP) \citep{garnelo2018conditional}. 
Instead of a latent variable, CNPs directly learn the predictive conditional distribution via a maximum likelihood meta-training procedure.  
Consequently they lack the ability to produce coherent function samples since each point is generated independently.
Most subsequent work has improved upon NPs (and CNPs) through architectural modifications such as attention \citep{kim2019attentive, kim2022neural}, convolutions \citep{gordon2019convolutional, foong2020meta}, mixtures \citep{wang2022moe}, equivariance \citep{kawano2021group}, and adaptation \citep{requeima2019fast}.  Comparatively fewer works have attempted to improve the distributional assumptions of the latent variable.
Two works have attempted to employ hierarchical \citep{wang2020doubly} and non-parametric \citep{flamstick} formulations to solve this problem.

\paragraph{Sum-Decomposition Networks}
The inference networks for NPs must have at least two properties.  The first is that they make no assumptions about the size of the context set.  The second is that the encoder be invariant to the ordering of context points.  A common way to satisfy these criteria is by having the encoder take the form of a sum-decomposition network \citep{edwards2016towards,zaheer2017deep}:

\begin{equation}
   \textstyle 
   \bar{\vr} = \frac{1}{N_c} \sum_{i=1}^{N_c} \vr_{c,i} \quad \textrm{with} \;
   \vr_{c,i} = \textrm{enc}_{\phi}^{r}(\vx_{c,i}, \vy_{c,i}), 
   \label{eq:mean_agg}
\end{equation}
where $\vr_i$ are datapoint-wise encodings given by a NN which are then aggregated. The aggregation operation is typically taken to be a simple average in NPs but other operators are valid as long as they are permutation-invariant.
Thus the amortisation goes one level further with parameter sharing across context points.
Finally, the aggregated representation $\bar{\vr}$ is passed to a further NN to give the variational parameters.
In the case of Gaussian posterior we have $\big(\widetilde{\vmu},\widetilde{\MSigma}\big) = \textrm{enc}_{\phi}^{z}(\bar{\vr})$.

\paragraph{Bayesian Context Aggregation}
\citet{volpp2020bayesian} propose a novel aggregation mechanism, derived by constructing a surrogate conditional latent variable (CLV) model in which the datapoint-wise encodings $\vr_i$ are interpreted as noisy observations of the underlying Gaussian latent variable.
Another encoder network then evaluates the observation noise for each datapoint $\vsigma_{c,i}^2 = \textrm{enc}_{\phi}^{\textrm{sigma}}(\vx_{c,i}, \vy_{c,i})$.
By choosing a Gaussian prior with mean $\vmu_0$ and variance $\vsigma_0^2$, Bayesian inference in this surrogate CLV results in the following aggregation mechanism:
\begin{align}
\begin{split}
    \widetilde{\vsigma}^2 &= \textstyle\sqr{ \vsigma_0^{-2} + \sum_{i=1}^{N_c} \vsigma_{c,i}^{-2} }^{-1}  \\
    \widetilde{\vmu} &= \textstyle\vmu_0 + \widetilde{\vsigma}^2 \circ \sum_{i=1}^{N_c} \rnd{ \vr_{c,i} - \vmu_0 } / \vsigma_{c,i}^2  \label{eq:bca_mean}
\end{split}
\end{align}
where $\va \circ \vb$ and $\va / \vb$ denote element-wise product and division respectively between vector $\va$ and $\vb$.
Aggregation operates directly in the latent-space which forgoes the need for a further NN.
Bayesian aggregation is a strict generalization of mean aggregation since the latter is recovered when a non-informative prior is used along with uniform observation variances.
\cref{eq:bca_mean} can be seen as re-weighting context points, with the weights given by $\vsigma_{c,i}^{-2}$.
This has some resemblance to self-attention mechanisms that have been adapted for neural processes \citep{kim2019attentive} however the key difference is that the weights are computed without consideration of other context points.

\section{Structured Inference Networks} \label{sec:sin}

We describe our approach that reframes individual context embeddings of the sum-decomposition architecture as factors in a probabilistic graphical model.
To motivate this, let us consider the case where the likelihood and prior in \cref{eq:post_true} are given by \emph{conjugate} exponential-family distributions.
The posterior can then be obtained analytically \citep{wainwright2008graphical}.
The prior can be expressed as
$p(\vz) = h(\vz) \exp \sqr{\ang{\MT(\vz), \veta_0} - A(\veta_0)}$,
where $\veta_0$ is the natural parameters, $\MT(\vz)$ is the sufficient statistics, $h(\vz)$ is the base measure and $A(\veta_0)$ is the log-partition function.
Due to conjugacy, the likelihood can be expressed in the same form as the prior
for some natural parameters $\veta_y(\vz)$ and sufficient statistics $\MT(\vy_{c,i})$,
\begin{align*}
    p(\vy_{c,i} | \vz) &= h(\vy_{c,i}) \exp \sqr{ \ang{\MT(\vy_{c,i}), \veta_y(\vz)} - A\rnd{\veta_y(\vz)}} \\
    &\propto \exp \Bigg( 
    {\underbrace{\begin{bmatrix} \veta_y(\vz) \\ -A\rnd{\veta_y(\vz)} \end{bmatrix}}_{=\MT(\vz)}}^\T%
    \underbrace{\begin{bmatrix} \MT(\vy_{c,i}) \\ 1 \end{bmatrix}}_{=\veta_{c,i}(\vy_{c,i})}
    \Bigg)
\end{align*}
where we have excluded the feature vector for simplicity.
Then the posterior distribution is,
\begin{equation}\textstyle
    p(\vz | \data) \propto h(\vz) \exp \sqr{ \Big\langle\MT(\vz), \veta_0 + \sum_{i=1}^N \veta_{c,i}(\vy_{c,i}) \Big\rangle }
    \label{eq:conjuagate_update}
\end{equation}
where it is clear that the posterior natural parameters are given by simply adding the \emph{sufficient statistics} of $\{\vy_{c,i}\}_{i=1}^{N_c}$ to the prior natural parameters.
The computation in \cref{eq:conjuagate_update} is strikingly similar to the sum-decomposition architecture in \cref{eq:mean_agg}.
There, the encodings $\textrm{enc}_{\phi}^{r}(\vx_{c,i}, \vy_{c,i})$ play a similar role to the sufficient statistics $\veta_{c,i}(\vy_{c,i})$ in that they are aggregated to obtain the contextual representation; but they are non-linear embeddings which could be much more expressive.

\subsection{Neural Sufficient Statistics}
Using the recipe for conjugacy in the exponential-family, we aim to construct a variational distribution of the same form but replacing the sufficient statistics by \emph{neural sufficient statistics}.
While training the end-to-end NP still requires variational inference (due to the NNs in the encoder and decoder), having this form will allow for efficient, conjugate updates to the distribution over the latent variables.
This is given as,
\begin{align}
\begin{split}
    &q_{\phi}(\vz | \data_c) \\
    &\quad \propto \textstyle\exp \sqr{ \Big\langle\MT(\vz), \veta_{\phi_{\textrm{PGM}}} + \sum_{i=1}^{N_c} \vf_{\phi_{\textrm{NN}}}(\vx_{c,i}, \vy_{c,i}) \Big\rangle } 
    \label{eq:neural_suff}
\end{split}
\end{align}
where $\vf_{\phi_\textrm{NN}}(\cdot)$ is a neural network for amortized and gradient-based 
construction of neural sufficient statistics \citep{wu2020amortized}
and $\veta_{\phi_{\textrm{PGM}}}$ is the prior natural parameters.
This is an instance of a structured inference network (SIN) \citep{lin2018variational} 
in which the variational distribution takes a factorized form consisting of the prior and $N$ \emph{factors} which are sometimes referred to as deep observational likelihoods \citep{johnson2016composing}.
This can be seen by rewriting \cref{eq:neural_suff} as,
\begin{align}
    q_{\phi}(\vz \mid \data_c) \mspace{-2mu} = \mspace{-2mu} \frac{1}{Z_c(\phi)} 
    &\underbrace{\Bigg[ \prod_{i=1}^{N_c} \exp \Big( \big\langle\MT(\vz), \vf_{\phi_{\textrm{NN}}}(\vx_{c,i}, \vy_{c,i}) \big\rangle \Big) \Bigg]}_{\textrm{NN factors}} \nonumber \\
    & \quad  \quad  \quad \quad  \quad  \quad \ \  \times \underbrace{\Bigg[ q(\vz ; \phi_{\textrm{PGM}}) \Bigg]}_{\textrm{prior}} \label{eq:sin}
\end{align}
where $Z_c(\phi)$ is the normalization constant.
The framework is also flexible to allow for the prior parameters to be fitted, $\phi := \{ \phi_{\textrm{NN}}, \phi_{\textrm{PGM}} \}$.
The neural sufficient statistics give rise to factors that can be conjugate to the prior by construction. 
Whilst this may seem like an arbitrary construction, in the specific case of Gaussianity, the optimal variational approximation decomposes into the prior and local Gaussian factors that approximate the likelihood \citep{nickisch2008approximations,opper2009variational}.

Furthermore, the variational lower bound resulting from using the structured inference network (\cref{eq:sin}) in \cref{eq:elbo} contains a term that resembles the entropy of the individual factors (see \cref{sup:sin_elbo} for further details).
This is further evidence that the factors are approximating the sufficient statistics of the likelihood; this is due to the link between statistical sufficiency and information-maximizing representations of the data \citep{chen2021neural}.

The structured inference network improves several aspects of the sum-decomposition network.
The first improvement is the variational parameters are now computed directly from the context embeddings.
This eliminates the need for an additional neural network reducing the total number of encoder parameters.
The second improvement is the introduction of an explicit prior distribution whose parameters are aggregated along with the context embeddings.
The third improvement is the aggregation mechanism is determined directly from the exponential-family parameterization.
Natural parameterization, as shown in \cref{eq:neural_suff}, leads to sum-pooling which is equivalent to mean-pooling in \cref{eq:mean_agg} within a constant of proportionality.
However, expectation parameterization leads to weighted aggregation mechanisms as we demonstrate below for two cases with Gaussian prior and Mixture of Gaussian prior.

\subsection{Bayesian context aggregation as SIN with Gaussian assumptions}\label{sec:bca_sin}
For a Gaussian prior, the resulting conjugate exponential-family distribution is also Gaussian.
By substituting this in \cref{eq:sin} with $\MT(\vz) := \{ \vz, \vz\vz^\T \}$ for the factors we obtain,
\begin{align*}\textstyle
    q_{\phi}(\vz \mid \data_c) &= \textstyle\frac{1}{Z_c(\phi)} \sqr{\prod_{i=1}^{N_c} \N(\vz \mid \vm_{c,i}, \MV_{c,i})} \\
    & \qquad \qquad \qquad \times \Big[\N(\vz \mid \vmu_0, \MSigma_0)\Big]
\end{align*}
where $\big( \vm_{c,i}, \MV_{c,i} \big) \leftarrow \vf_{\phi_{\textrm{NN}}}(\vx_{c,i}, \vy_{c,i})$ are the moment parameters of the factors (mean and variance) evaluated by the recognition network and prior moments $\phi_{\textrm{PGM}} := \{ \vmu_0, \MSigma_0 \}$.
Then the variational distribution is also Gaussian $q_{\phi}(\vz \mid \data_c) = \N(\vz ; \widetilde{\vmu}, \widetilde{\MSigma})$ with posterior moments given by,
\begin{align}
\begin{split}
    \widetilde{\MSigma}^{-1} &= \textstyle\sum_{i=1}^{N_c} \MV_{c,i}^{-1} + \MSigma_0^{-1} \\ %
    \widetilde{\vmu} &= \widetilde{\MSigma} \rnd{\textstyle\sum_{i=1}^{N_c} \MV_{c,i}^{-1} \vm_{c,i} + \MSigma_0^{-1} \vmu_0}. \label{eq:gauss_mean}
\end{split}
\end{align}
The normalization constant $Z_c(\phi)$ is also available in closed-form.
We also assume diagonal covariance matrices throughout thereby ensuring no expensive matrix operations are performed for aggregation.
This Gaussian-based procedure is equivalent to the previously proposed Bayesian aggregation mechanism \citep{volpp2020bayesian}.
This can be seen by a straightforward manipulation of \cref{eq:gauss_mean} to give the incremental update form of \cref{eq:bca_mean} (see \cref{sup:bca_equivalence} for proof).

\subsection{Mixture Bayesian Aggregation}
We now consider a more expressive prior distribution, the mixture of Gaussian (MoG) prior, which despite being a conditionally-conjugate exponential-family distribution, results in closed-form updates when the factors are chosen to be Gaussian,
\begin{align*}
    q_{\phi}(\vz \mid \data_c) &= \textstyle\frac{1}{Z_c(\phi)} \sqr{\textstyle\prod_{i=1}^{N_c} \N(\vz \mid \vm_{c,i}, \MV_{c,i})} \\
    & \qquad \qquad \qquad \times \sqr{ \textstyle\sum_{k=1}^K \pi_k \N(\vz \mid \vmu_k, \MSigma_k) }. \nonumber %
\end{align*}
The factors' moment parameters are evaluated in the same way as \cref{sec:bca_sin} and the prior parameters are given by $\phi_{\textrm{PGM}} := \{ \pi_k, \vmu_k, \MSigma_k \}_{k=1}^K$ where $\sum_{k=1}^K \pi_k = 1$.
Then the variational distribution also takes a MoG form $q_{\phi}(\vz \mid \data_c) = \sum_{k=1}^K \widetilde{\pi}_k \N(\vz \mid \widetilde{\vmu}_k, \widetilde{\MSigma}_k)$ with posterior parameters given by,
\begin{align}
\begin{split}
    \widetilde{\MSigma}_k^{-1} &= \textstyle\sum_{i=1}^{N_c} \MV_{c,i}^{-1} + \MSigma_k^{-1} \\
    \widetilde{\vmu}_k &= \widetilde{\MSigma}_k \rnd{\textstyle\sum_{i=1}^{N_c} \MV_{c,i}^{-1} \vm_{c,i} + \MSigma_k^{-1} \vmu_k} \\
    \widetilde{\pi}_k &= \pi_k \; C_k \; / \; Z_c \\
\end{split}
\end{align}
for $k=1,\ldots,K$ where $Z_c = \sum_{k=1}^K \widetilde{\pi}_k$.
The updates for the mean and variance of each component Gaussian takes an identical form to \cref{eq:gauss_mean}.
The update for the mixing proportions requires evaluation of the normalization constant $C_k$ for each Gaussian component which is stated in \cref{sup:mog}.
A MoG prior may be a beneficial modelling assumption if we expect the data to arise from multiple sources.
We refer to this approach as \emph{mixture} Bayesian Aggregation (mBA), a generalization of BA that is recovered when the number of mixture components $K$ is set to 1.

\section{Beyond Conjugacy} \label{sec:sin_nonconjugate}
The conjugate case is attractive due to its analytic properties, but it may be a poor modelling assumption for real-world data.
By relaxing the need for conjugacy between the factors and the prior, we can allow for a wider range of distributional assumptions. %
Yet non-conjugacy implies the posterior can no longer be expressed analytically since the evidence $Z_c(\phi)$ is intractable.
We can instead form a lower bound on the evidence by introducing an approximating distribution $\tilde{q}$,
\begin{equation}
    \log Z_c(\phi) \geq \E_{\tilde{q}(\vz)} \sqr{\log q_\phi(\vz, \data_c)} + \mathcal{H}\rnd{\tilde{q}(\vz)}
    \label{eq:non_conj_elbo}
\end{equation}
with joint distribution $q_\phi(\vz, \data_c)$ corresponding to \cref{eq:sin} and $\mathcal{H}(q) = \E_q\sqr{-\log q(\vz)}$ is the entropy.
The variational posterior now depends implicitly on the neural sufficient statistics and prior parameters.
Whilst this approach would be generally applicable, 
without further assumptions every forward pass through the NP would require solving the stochastic optimization problem that is maximizing \cref{eq:non_conj_elbo}.
This closely resembles a recently proposed aggregation mechanism for set embedding termed equilibrium aggregation \citep{bartunov2022equilibrium} which also considers an optimization-based formulation.
They show that under certain conditions (\eg choice of initialization, regularization strength \etc), convergent dynamics are observed with a small number of gradient-descent steps.

These techniques could also be adapted here, but instead we consider certain restrictions that lead to a more tractable approach while still allowing for expressive PGMs.
We start by restricting $\tilde{q}$ to be a mean-field distribution \citep{jordan1999introduction}.
Considering a certain partition of $\vz$ into $M$ disjoint groups over which $\tilde{q}$ factorises, we have $\tilde{q}(\vz) = \prod_{j=1}^M \tilde{q}_j(\vz_j)$.
Now the stationary point $\tilde{q}^*(\vz)$ of \cref{eq:non_conj_elbo} satisfies (\citet{bishop2006pattern}, Eq.~10.9),
\begin{equation}
    \log \tilde{q}_j^*(\vz_j) = \E_{\tilde{q}^*_{/j}(\vz_{/j})} \sqr{\log q_\phi\rnd{\vz,\data_c}} + \textrm{cnst.}
    \label{eq:mean-field}
\end{equation}
where $\vz_{/j}$ indicates all variables excluding the $j$\textsuperscript{th} group using which the expectation of the log joint is taken with respect to.
We further suppose that the factors and prior specify a conditionally-conjugate exponential-family system and that each factor in $\tilde{q}(\vz)$ belongs to the same exponential-family as the complete conditional distribution in $q_\phi(\vz, \data_c)$.
Then the expectation in \cref{eq:mean-field} takes a closed-form expression resulting in tractable updates for each factor of the mean-field posterior.
We can iteratively optimize each $\vz_j$ whilst holding the others fixed using \cref{eq:mean-field}. %
This is often referred to as 
coordinate ascent variational inference (CAVI) \citep{blei2017variational} which is a special case of variational message passing \citep{winn2005variational}.
A closely related approach is differentiable EM (DIEM) \citep{kim2021differentiable} that frames set embedding as maximum-a-posteriori estimation.
This performs expectation-maximization (EM) updates and can be viewed as a special case of our method.

\begin{table*}[t]
\centering
\caption{\emph{mixture Bayesian Aggregation (mBA)}. We observe a boost in performance over BA with increasing number of components (K). We also exceed the performance of NP with Self-Attention (NP+SA) on OOD data. This is demonstrated on the task of 2D image completion (EMNIST).}
\label{tab:mog_emnist}
\resizebox{.99\linewidth}{!}{%
\input{tables/mog_emnist}
}
\end{table*}
\subsection{Robust Bayesian Aggregation}
We now consider a specific instance of a conditionally-conjugate exponential-family system in which the coordinate-ascent updates arising from \cref{eq:mean-field} give rise to a novel, weighted aggregation mechanism.
This extends the all-Gaussian assumptions of \cref{sec:bca_sin} by introducing a Gamma prior over the precision of each Gaussian factor, whose marginal form is a heavy-tailed Student-\emph{t} distribution, as well as a hierarchical prior.
This is adapted from previous work on robust Bayesian interpolation \citep{tipping2005variational} which demonstrated robustness to outliers and corruptions in the targets with this model specification.
The probabilistic model is,
\begin{align*}
    q_\phi(\vz,\alpha,\vbeta) &= \sqr{\textstyle\prod_{i=1}^{N_c} \N(\vz | \vm_{c,i}, \beta_i^{-1} \MV_{c,i})} \\
    & \qquad \qquad  \times \sqr{ q(\vz | \alpha) q(\alpha) \textstyle\prod_{i=1}^{N_c} q(\beta_i) }
\end{align*}
with $q(\vz | \alpha) = \N(\vz | \vzero, \alpha^{-1} \MI)$, $q(\alpha) = \G(\alpha | a_0, b_0)$ and $q(\beta_i) = \G(\beta_i | c_0, c_0)$.
The factors' moment parameters are evaluated as before and $\phi_{\textrm{PGM}} = \{a_0,b_0,c_0\}$.
Each factor and $q(\beta_i)$ together can be viewed as the hierarchical form of the Student-\emph{t} distribution,
\newcommand{\studentt}{\mathcal{T}}   %
\begin{align*}
    \studentt(\vz | \vm_{c,i}, \MV_{c,i}, c_0) &= 
    \textstyle\int \N(\vz \mid \vm_{c,i}, \beta_i^{-1} \MV_{c,i}) \\
    & \qquad \qquad \cdot \G(\beta_i | c_0, c_0) \; d\beta_i.
\end{align*}
The marginal form would render the updates arising from \cref{eq:mean-field} intractable hence we restrict ourselves to the joint specification.
Next we introduce a mean-field distribution with a corresponding factorization and functional form to the prior, $\tilde{q}(\vz, \alpha, \vbeta) = \tilde{q}(\vz) \tilde{q}(\alpha) \tilde{q}(\vbeta)$.
where $\tilde{q}(\vz) = \N(\vz | \widetilde{\vmu}, \widetilde{\MSigma})$, $\tilde{q}(\alpha) = \G(\alpha | \widetilde{a}, \widetilde{b})$ and $\tilde{q}(\vbeta) = \prod_{i=1}^{N_c} \G(\beta_i | \widetilde{c}, \widetilde{d}_i)$.
Plugging this into \cref{eq:mean-field}, we can derive the following updates for the parameters of the mean-field posterior,
\begin{align}
    \widetilde{\MSigma}^{-1} &= \textstyle\sum_{i=1}^{N_c} \E[\beta_i] \MV_{c,i}^{-1} + \E[\alpha] \MI \label{eq:gauss_var_iter} \\ 
    \widetilde{\vmu} &= \widetilde{\MSigma} \textstyle\sum_{i=1}^{N_c} \E[\beta_i] \MV_{c,i}^{-1} \vm_{c,i}, \label{eq:gauss_mean_iter} \\
    \widetilde{a} &= a_0 + \textstyle\frac{D}{2} \\
    \widetilde{b} &= b_0 + \textstyle\frac{1}{2} \E[\vz^\T \vz] \label{eq:gamma_b} \\
    \widetilde{c} &= c_0 + \textstyle\frac{D}{2} \\
    \widetilde{d}_i &= c_0 + \textstyle\frac{1}{2} \Big( 
    \vm_{c,i}^\T \MV_{c,i}^{-1} \vm_{c,i} - 2 \vm_{c,i}^\T \MV_{c,i}^{-1} \E[\vz]  \label{eq:gm_noise_d}  \\
    & \hspace{5em} + \tr(\MV_{c,i}^{-1} \E[\vz \vz^\T]) \notag
    \Big)
\end{align}
with expectations given by $\E[\vz] = \widetilde{\vmu}$, $\E[\vz \vz^\T] = \widetilde{\vmu}\widetilde{\vmu}^\T + \widetilde{\MSigma}$, $ \E[\alpha] = \textstyle\sfrac{\widetilde{a}}{\widetilde{b}}$ and $\E[\beta_i] = \textstyle\sfrac{\widetilde{c}}{\widetilde{d}_i}$. 
We iterate through the updates in the order presented above.
At convergence or after a fixed number of iterations, we only keep the posterior over $\vz$ which is passed to the NP decoder.

The new procedure is an extension of BA.
This can be seen by initializing the Gamma parameters to $\widetilde{a} = \widetilde{b} = \widetilde{c} = \widetilde{d}_i = 1$ and running a single step of \cref{eq:gauss_var_iter,eq:gauss_mean_iter}, then BA is recovered (when standard normal prior is used).
However, compared with BA, \cref{eq:gauss_var_iter} presents an alternate approach to updating the prior precision which now depends on the aggregated contextual representation through \cref{eq:gamma_b}.
This adaptive mechanism resembles that of a data-dependent prior \citep{tipping2001sparse}.

Another crucial difference in \cref{eq:gauss_var_iter,eq:gauss_mean_iter} is that the neural sufficient statistics (in natural parameterization) are now weighted by the 1\textsuperscript{st} moment of the variational noise distribution.
The context-dependent variability in this term is given by \cref{eq:gm_noise_d} which is evaluated by using the individual context embedding as well as the aggregated contextual representation given by the moments of $\tilde{q}(\vz)$.
Viewed as a weighted aggregation scheme, this has close resemblance to self-attention which evaluates a similarity function (\eg Euclidean norm, dot-product \etc) between all pairs of context points.
In the case of corruptions to the context set, such outliers can be downweighted, improving the robustness of neural processes.
We refer to our approach as \emph{robust} Bayesian aggregation (rBA).

We run the coordinate-wise updates for a fixed number of steps and observe that in practice convergent dynamics appear within a few steps as shown in \cref{fig:rr_elbo}.
We note that all operations are fully-differentiable and we backprop through the \emph{unrolled} steps for gradient-based learning.
It may be possible to incorporate implicit differentiation techniques, for instance as used in Deep Equilibrium Models \citep{bai2019deep}, to improve the efficiency and reduce the memory of the proposed method.

\begin{figure}[t]
    \centering
    \includegraphics[width=0.85\linewidth]{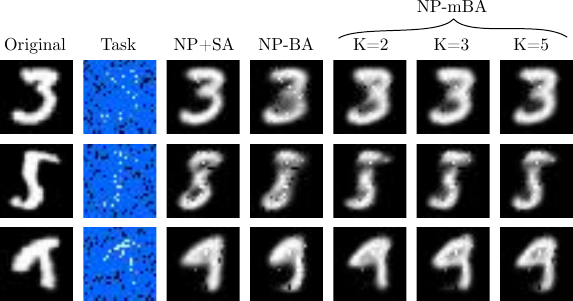}
    \caption{Completed images on EMNIST (0-9) from 100 context points (``task'').}
    \label{fig:mBA_emnist}
\end{figure}

\section{Experiments}
\begin{figure*}[t]
    \begin{subfigure}{\linewidth}
        \centering
        \begin{subfigure}{.49\linewidth}
            \centering
            \includegraphics[width=.97\linewidth]{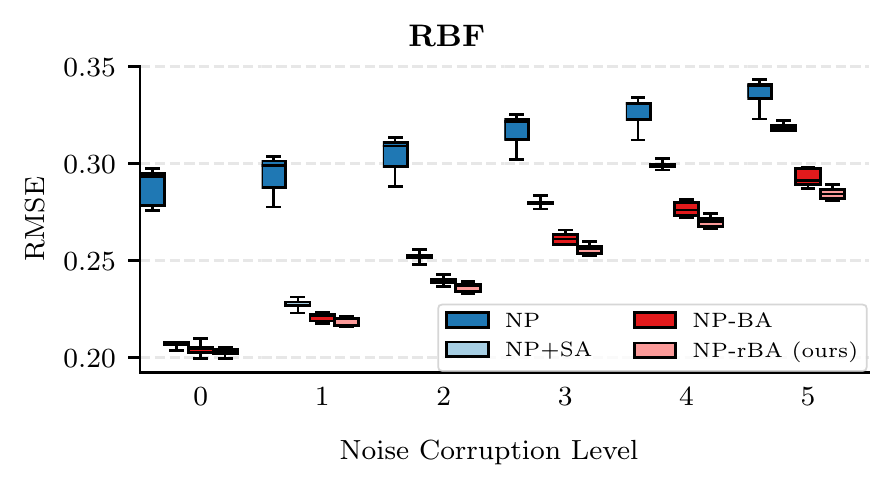}
        \end{subfigure}
        \hfill
        \begin{subfigure}{.49\linewidth}
            \centering
            \includegraphics[width=.97\linewidth]{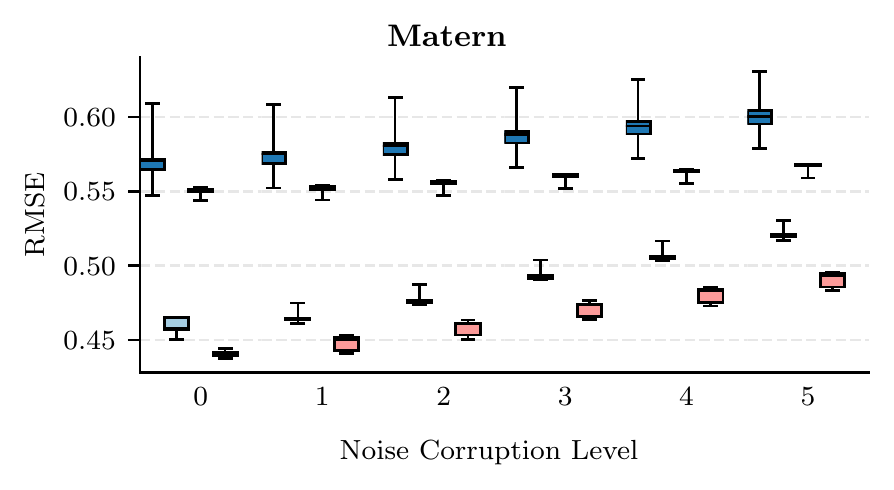}
        \end{subfigure}
        \caption{1D regression with functional samples from GP kernel.}
    \end{subfigure}\\
    \begin{subfigure}{\linewidth}
        \centering
        \begin{subfigure}{.49\linewidth}
            \centering
            \includegraphics[width=.97\linewidth]{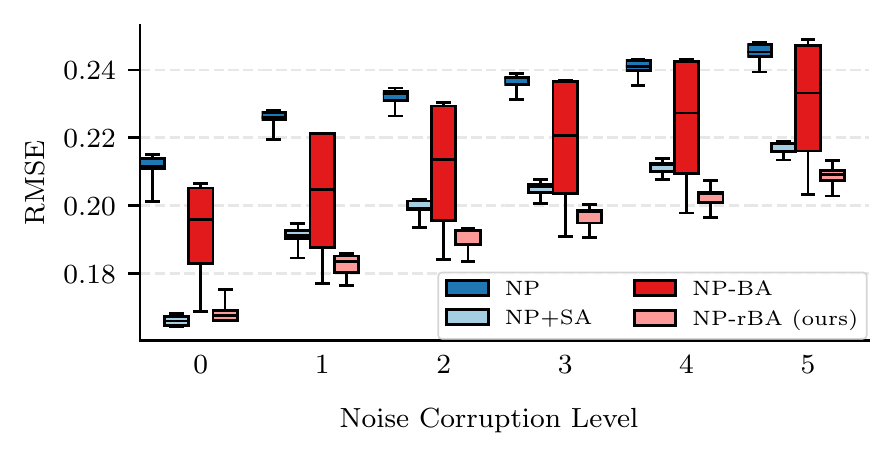}
        \end{subfigure}
        \hfill
        \begin{subfigure}{.49\linewidth}
            \centering
            \includegraphics[width=.97\linewidth]{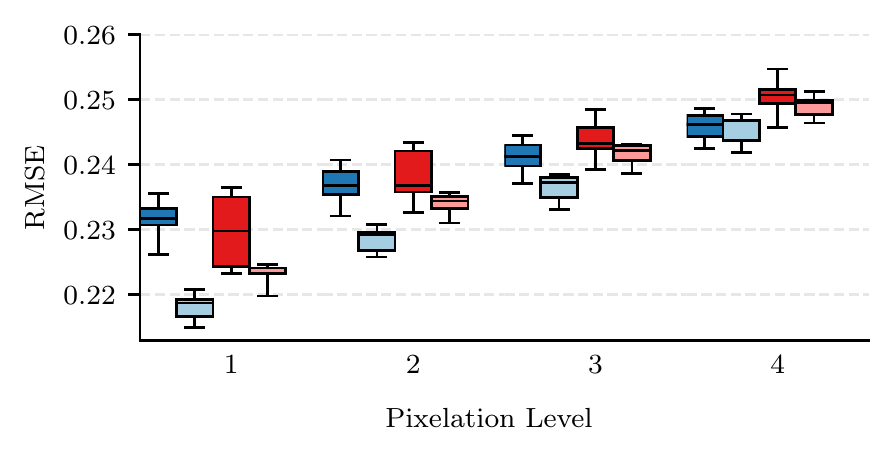}
        \end{subfigure}
        \caption{2D image completion on EMNIST dataset.}
    \end{subfigure}
    \caption{\emph{robust Bayesian Aggregation (rBA).} We demonstrate gains to test-time robustness in the presence of corruptions to the context set as compared with BA. In (a), context set function values are perturbed by heavy-tail noise of varying magnitude. In (b), we consider common image noise perturbations (left) and image pixelation (right), of varying intensity.
    }
    \label{fig:rBA_corruption}
\end{figure*}
We conduct experiments to assess the performance of our mixture and robust variants\footnote{For a reference implementation see: \url{https://github.com/dvtailor/np-structured-inference}.} of Bayesian Aggregation (BA) against vanilla BA \citep{volpp2020bayesian}.
We also benchmark against NPs with regular encoder architecture (NP) \citep{garnelo2018neural} and an NP with self-attention module (NP+SA) \citep{kim2019attentive}.
Similar to \citet{volpp2020bayesian}, we only evaluate on NP-based models with a single latent path (i.e. no deterministic path).
We do not consider encoder architectures that evaluate task-specific contextual representations such as Attentive NP \citep{kim2019attentive} and leave the combination of BA variants with cross-attention style mechanisms to future work.
The decoder network architecture is fixed across all models such that any performance differences can be attributed solely to the encoder architecture.
To reduce the effects of differences in network architecture between BA and non-BA encoders, we keep the BA encoder size close to but less that of the non-BA architectures.
Each model is trained with 5 different seeds and we report the mean and standard deviations from these runs.
Please refer to \cref{sup:exp} for further details on the model specification and training configuration.

We consider the following tasks,
\begin{itemize}[leftmargin=*]
    \item \textbf{1-D Regression} We train on functional samples drawn from Gaussian Process (GP) priors with RBF and Matern-5/2 kernel. The specification of the RBF kernel is taken from \citet{lee2020bootstrapping} where both the lengthscale and output variance is varied (GP hyperparameters). The matern-5/2 kernel is taken from \citet{gordon2019convolutional} where the hyperparameters are fixed.
    \item \textbf{Image completion} A subset of pixels for a given image is presented with the task to predict the remaining pixels. Therefore, we can understand the inputs to be coordinates of each pixel and the pixel intensities representing the targets (i.e. 2-D regression). We consider the EMNIST dataset \citep{cohen2017emnist} -- a dataset of handwritten digits and letters comprising of grayscale 28x28 images. During training, we restrict images to the first 10 classes. We evaluate on two settings: in-distribution data (held-out images from the first 10 classes) and out-of-distribution (OOD) data (images from the remaining 37 classes).
\end{itemize}

\newcommand{\rulesep}{\unskip\ \vrule\ }
\begin{figure*}[t]
    \centering
    \begin{subfigure}{.49\linewidth}
        \begin{subfigure}{.49\linewidth}
            \centering
            \includegraphics[width=\linewidth]{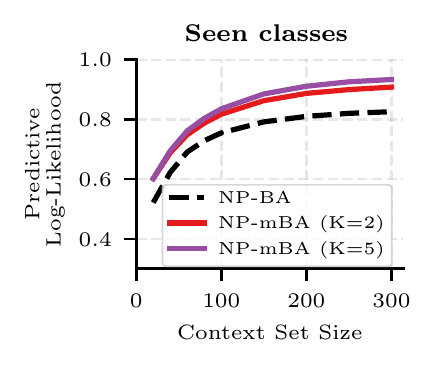}
        \end{subfigure}
        \hfill 
        \begin{subfigure}{.49\linewidth}
            \centering
            \includegraphics[width=\linewidth]{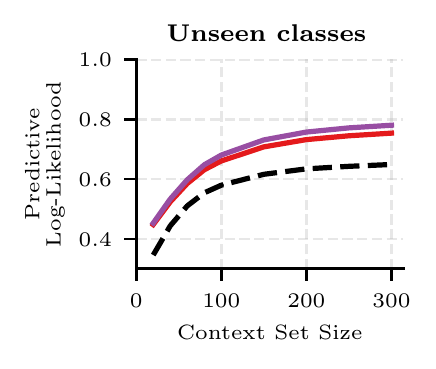}
        \end{subfigure}
        \caption{Boost with increasing mixture components}
        \label{fig:mog_emnist_context}
    \end{subfigure}
    \rulesep
    \begin{subfigure}{.49\linewidth}
        \begin{subfigure}{.49\linewidth}
            \centering
            \includegraphics[width=\linewidth]{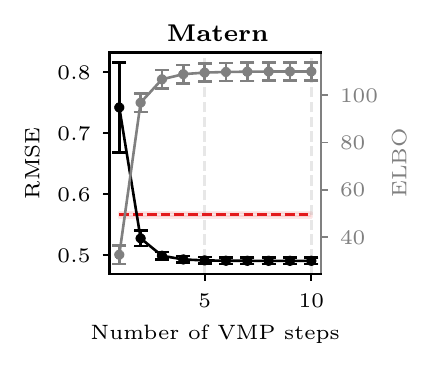}
        \end{subfigure}
        \hfill 
        \begin{subfigure}{.49\linewidth}
            \centering
            \includegraphics[width=\linewidth]{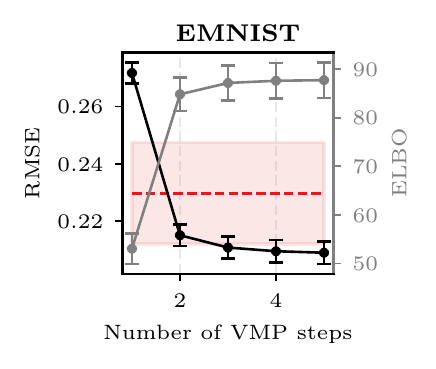}
        \end{subfigure}
        \caption{Boost with increasing message-passing steps}
        \label{fig:rr_elbo}
    \end{subfigure}
    \caption{%
    (a) Across a range of context set sizes, there is a gain in performance with increasing mixture components over BA. This is demonstrated on the task of 2D image completion (error bars are omitted for visual clarity).
    (b) By increasing the number of message-passing (VMP) steps, we observe a monotonic improvement in both the accuracy of the target function  and on the lower bound to the evidence of the PGM. With just 2 steps, there is a gain over vanilla BA (\protect\tikz[baseline=-0.5ex,inner sep=0pt]{\protect\draw[line width=1.5pt, red, dashed] (0,0) -- ++(0.3,0);}). This is shown for test-time corruption to the context set of the highest intensity. During training, a fixed number of steps is used (10 and 5 for Matern and EMNIST respectively).}
\end{figure*}

\subsection{Mixture Bayesian Aggregation} \label{sec:exp_mba}
We initialize the mixture prior in the PGM as follows: 
the prior mixing proportions are set to uniform, the prior variance is set to 1 and the prior means are sampled from a zero-mean Gaussian with standard deviation $0.1$. The latter setting is to encourage diversity in the posterior mixture components and prevent collapse to a unimodal Gaussian.
The task of image completion on the EMNIST dataset is considered.
\cref{tab:mog_emnist} clearly demonstrates the utility of additional mixing components with improved function modelling as the number of components is increased from 2 to 5 (see \cref{fig:mBA_emnist}).
This is also shown in dependence of the context set size in \cref{fig:mog_emnist_context}.
The reconstruction error is also indicated (``context'').
On OOD data, NP-mBA even exceeds the performance of NP with Self-Attention (NP+SA) for all numbers of components considered.
This is despite NP+SA having a considerably larger number of parameters (see \cref{tab:runtime_image}). We also observe little change in the run-time when the number of components is increased.

\subsection{Robust Bayesian Aggregation} \label{sec:exp_rba}
\begin{figure}[t]
    \centering
    \includegraphics[width=0.85\linewidth]{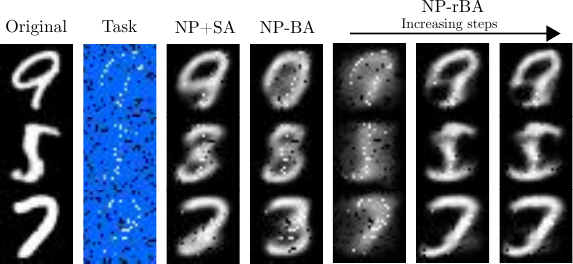}
    \caption{Completed images on EMNIST (0-9) from 100 context points (``task'') that have been corrupted by Gaussian noise.}
    \label{fig:rBA_emnist}
\end{figure}

We demonstrate the improved robustness of our robust Bayesian Aggregation (NP-rBA) to corruptions in the context sets at test-time (see \cref{fig:rBA_emnist}).
For the 1D regression task, we extend the model-data mismatch setting from \citet{lee2020bootstrapping} where the function values are corrupted by Student-\emph{t} noise of increasing magnitude, $\varepsilon \sim \gamma \cdot \mathcal{T}(2.1)$ with $\gamma \in [0.05,0.08,0.11,0.13,0.15]$.
For the image completion setting, we only consider the in-distribution data and perform noise corruption and pixelation as specified in \citep{hendrycksbenchmarking}.
For the noise corruption, we average over four different types of noise, namely Gaussian, Shot, Poisson and Impulse noise, of increasing severity.
Pixelation involves downsampling the image to different reduced resolutions and then upsampling back to the original resolution.
For our NP-rBA, we run the message passing algorithm for 5 steps (at both train and test time) except for Matern where we run for 10 steps.

In \cref{fig:rBA_corruption}, we observe a consistent gain over BA across the different tasks and corruptions.
With the exception of corruption by pixelation, NP-rBA also shows improvement over NP with Self-Attention (NP+SA).
\cref{fig:rr_elbo} demonstrates speed-accuracy trade-off where increasing the number of steps at test-time leads to monotonic improvement in performance.

\begin{table}[t]
\centering
\caption{Number of model parameters and training time (s). Run-time is measured on the image completion experiment on a GTX 1080, for a single epoch (batch size 100). NP-mBA is run with 5 components and NP-rBA with 5 message-passing steps.}
\input{tables/runtime_image}
\label{tab:runtime_image}
\end{table}

\section{Conclusion}
We propose structured inference networks for context aggregation in Neural Processes (NPs). This change is attractive for several reasons: (i) the local encodings now have a clear interpretation as neural sufficient statistics, (ii) the aggregation step is predetermined by and follows from the probabilistic assumptions and inference routine, and (iii) structured priors are straightforward to incorporate.
We show that an existing context aggregation mechanism, Bayesian Aggregation (BA), is recovered by imposing Gaussianity assumptions in the PGM.

By imposing different modelling assumptions, we demonstrate alternative context aggregation mechanisms can be derived.
In particular, we consider, (1) Mixture of Gaussian prior and (2) Student-\emph{t} assumptions, which give rise to two novel variants of BA, namely \emph{mixture} and \emph{robust} Bayesian Aggregation.
We demonstrate improvements to the functional modelling and test-time robustness of NPs without any increase in the parameterization of the encoder.

For future work, we look to consider more general PGM structures, \eg temporal transition structure for modelling time-series data %
or even causal structure via directed acyclic graphs (DAGs).
It is worth reiterating that there is effectively no systematic limitation to our framework since, if the user deems there to be too much computational overhead in using an expressive PGM, the assumptions can be simplified (\eg Gaussianity) thereby recovering existing strategies.

\begin{contributions} %
List of Authors: Dharmesh Tailor (D.T.), Mohammad Emtiyaz Khan (M.E.K.), Eric Nalisnick (E.N.).

D.T. and E.N. conceived the original idea. This was then discussed with M.E.K. D.T. derived and implemented the new aggregation strategies and ran the experiments, with regular feedback from E.N. D.T. wrote the first draft of the paper, after which all authors contributed to writing.
\end{contributions}

\begin{acknowledgements}
    We would like to thank Qi Wang (University of Amsterdam) for his helpful feedback and discussions.
\end{acknowledgements}

\bibliography{refs}

\newpage
\appendix
\onecolumn
\thispagestyle{empty}

\newcommand{\toptitlebar}{
  \hrule height 4pt
  \vskip 0.25in
  \vskip -\parskip%
}
\newcommand{\bottomtitlebar}{
  \vskip 0.29in
  \vskip -\parskip
  \hrule height 1pt
  \vskip 0.09in%
}

\toptitlebar
{\centering
\Large\bfseries Exploiting Inferential Structure in Neural Processes\\(Supplementary Material)\par}
\bottomtitlebar

\section{Derivations}
\subsection{Variational lower bound with structured inference network} \label{sup:sin_elbo}
For the conjugate case in \cref{sec:sin}, we show the ELBO in \cref{eq:elbo} (for a single task and dropping task indices for clarity) simplifies after substitution of the structured inference network (\cref{eq:sin}).
We denote the factor by $q(\vz \mid \vf_{\phi_{\textrm{NN}}}(\data_c^{(i)}) ) := \exp \rnd{ \big\langle\MT(\vz), \vf_{\phi_{\textrm{NN}}}(\vx_{c,i}, \vy_{c,i}) \big\rangle}$ (and analogous expression for each target point):
\begin{align}
    &\log p(\data_t \mid \data_c) \\
    &\phantom{{}=1}= \log \int_{\vz} p(\data_t \mid \vz) \; p(\vz \mid \data_c) \\
    &\phantom{{}=1}= \log \int_{\vz} q_{\phi}(\vz \mid \data_c \cup \data_t) \frac{p(\data_t \mid \vz) \; p(\vz \mid \data_c)}{q_{\phi}(\vz \mid \data_c \cup \data_t)} \\
    &\phantom{{}=1}\geq \E_{q_{\phi}(\vz \mid \data_c \cup \data_t)} \sqr{ \log \frac{p(\data_t \mid \vz) \; p(\vz \mid \data_c)}{q_{\phi}(\vz \mid \data_c \cup \data_t)} } \\
    &\phantom{{}=1}\approx \E_{q} \sqr{ \log \frac{p(\data_t \mid \vz) \;  q_{\phi}(\vz \mid \data_c)}{q_{\phi}(\vz \mid \data_c \cup \data_t)} } \\
    &\phantom{{}=1}= \E_q \sqr{ \log p(\data_t \mid \vz) }
    + \E_q \bigg[ \log
    \frac{%
    \cancel{\prod_{i=1}^{N_c} q(\vz \mid \vf_{\phi_{\textrm{NN}}}(\data_c^{(i)}) )} \; 
    }%
    {%
    \cancel{\prod_{i=1}^{N_c} q(\vz \mid \vf_{\phi_{\textrm{NN}}}(\data_c^{(i)}) )} 
    \prod_{i=1}^{N_t} q(\vz \mid \vf_{\phi_{\textrm{NN}}}(\data_t^{(i)}) )
    }%
    \frac{\cancel{q(\vz ; \phi_{\textrm{PGM}})} \; Z_{c,t}(\phi)}{\cancel{q(\vz ; \phi_{\textrm{PGM}})} \; Z_c(\phi)} \bigg] \\
    &\phantom{{}=1}= \E_q \sqr{ \log p(\data_t \mid \vz)}
    - \sum_{i=1}^{N_t} \textcolor{blue}{\E_q \sqr{ \log q(\vz \mid \vf_{\phi_{\textrm{NN}}}(\data_t^{(i)}) ) } }
    + \log Z_{c,t}(\phi) - \log Z_{c}(\phi)
\end{align}
where the 2\textsuperscript{nd} term resembles the entropy on the individual factors (shown in blue).

\subsection{EQUIVALENCE TO BAYESIAN AGGREGATION MEAN UPDATE EQUATION}\label{sup:bca_equivalence}
We show the posterior mean in \cref{eq:gauss_mean} can be expressed in the incremental form stated in \cite{volpp2020bayesian} (\cref{eq:bca_mean}):
\begingroup
\allowdisplaybreaks
\begin{align}
    \widetilde{\vmu} &= \widetilde{\MSigma} \rnd{\sum_{i=1}^{N_c} \MV_{c,i}^{-1} \vm_{c,i} + \MSigma_0^{-1} \vmu_0} \\
    &= \widetilde{\MSigma} \rnd{\sum_{i=1}^{N_c} \MV_{c,i}^{-1} \vm_{c,i} + \widetilde{\MSigma}^{-1} \vmu_0 - \sum_{i=1}^{N_c} \MV_{c,i}^{-1} \vmu_0 } \\
    &= \vmu_0 + \widetilde{\MSigma} \sum_{i=1}^{N_c} \MV_{c,i}^{-1} \rnd{\vm_{c,i} - \vmu_0}
\end{align}
\endgroup

\subsection{MIXTURE OF GAUSSIAN PRIOR NORMALIZATION CONSTANT}\label{sup:mog}
\begin{equation}
    C_k = (2\pi)^{-\frac{DN}{2}} \prod_{i=1}^{N_c} \det(\MV_{c,i})^{-\frac{1}{2}} \rnd{\frac{\det(\MSigma_k)}{\det(\widetilde{\MSigma}_k)}}^{-\frac{1}{2}}
    \exp\Bigg\{-\frac{1}{2} \Bigg( \sum_{i=1}^{N_c} \vm_{c,i}^\T \MV_{c,i}^{-1} \vm_{c,i} + \vmu_k^\T \MSigma_k^{-1} \vmu_k
    - \widetilde{\vmu}_k^\T \widetilde{\MSigma}_k^{-1} \widetilde{\vmu}_k \Bigg) \Bigg\}
\end{equation}

\section{EXPERIMENTAL DETAILS}\label{sup:exp}
The implementation for robust and mixture Bayesian aggregation is adapted from the implementation of BA\footnote{\url{https://github.com/boschresearch/bayesian-context-aggregation}} and the other baselines are taken from the codebase of Bootstrapped Neural Process\footnote{\url{https://github.com/juho-lee/bnp}} (with modifications elaborated in the appendix) \citep{lee2020bootstrapping}.

For our proposed mixture Bayesian Aggregation, we use the \texttt{MixtureOfDiagNormals} implementation from the \texttt{Pyro} package.
However due to numerical issues with using 32-bit floating-point precision with this implementation, we disable gradient flow through the categorical distribution.

\subsection{1-D regression}
We consider the following kernels:
\begin{enumerate}[leftmargin=8em,itemsep=0.5em]
    \item[RBF:]
        baseline used in \citet{lee2020bootstrapping}
        \[
            k(x, x') = s^2 \exp \rnd{ - \frac{(x-x')^2}{2 \ell^2} }
        \]
        with $s \sim \U[0.1, 1.0)$ and $\ell \sim \U[0.1,0.6)$;
    \item[Mat\'ern--$\frac52$:]
        baseline used in \citet{gordon2019convolutional}
        \[
            k(x, x') = \left(1 + \sqrt{5} d  + \frac53 d^2\right) \exp\left(-\sqrt{5} d \vphantom{\frac52}\right)
        \]
        with $d = 4|x - x'|$.
\end{enumerate}

Following \citet{lee2020bootstrapping}, the inputs of the context and target sets are sampled according to $x \sim \U(-2,2)$.
The sizes of the context and target sets are sampled according to $N_c \sim \U(3, 47)$ and $N_t \sim \U(3, 50-N_c)$.
In the evaluation phase, 5000 tasks are drawn identically to the data generating process for training.

\subsection{Image completion}
Image completion is formulated as a regression problem where pixel coordinates are transformed to $[-1,1]$ and pixel intensities are rescaled to $[-0.5,0.5]$ following \citet{lee2020bootstrapping}.
A single image constitutes a task.
During training, images are restricted to the first 10 classes, with the size of the context and target sets are sampled according to 
$N_c \sim \U(3,197)$ and $N_t \sim \U(3,200-N_c)$.
For the in-distribution setting, we evaluate on a different set of images but restricted also to the first 10 classes. %
For \cref{fig:mog_emnist_context} we sample the target sets according to $N_t \sim \U(3,500-N_c)$ (evaluation only).
For evaluation on the out-of-distribution setting, images are taken from the unseen classes 10-46.

\subsection{Model architectures}
\paragraph{Decoder architecture} Across all models, we keep the decoder architecture the same, with separate networks outputting the mean and standard deviation of the predictive distribution (following \citet{volpp2020bayesian}).
The networks have 128 hidden units and ReLU activation functions.
For the 1-D regression experiment, we use a 3-layer MLP and for the image completion experiment, a 4-layer MLP.
Following \citet{le2018empirical}, the standard deviation of the predictive distribution is processed using a lower-bounded softplus with a lower bound of $0.1$.

\paragraph{Encoder architecture}
For models with latent path, the latent dimensionality is set to 128. ReLU activation function for the hidden layers are used throughout. Unless otherwise stated, the hidden size is 128.
Here we state the architectures of the baselines with mean-pooling aggregation in the 1-D regression experiment:
\begin{enumerate}[leftmargin=8em,itemsep=0.5em]
    \item[NP:] This is adapted from \cite{garnelo2018neural} where the deterministic path is removed (as done in \citet{volpp2020bayesian}). 
    \item[NP+SA:] This incorporates (multi-head) self-attention into the encoder architecture of NP.
\end{enumerate}
Following \citet{le2018empirical}, we process the standard deviation of the latent variable using a lower-bounded sigmoid with a lower bound of $10^{-4}$.

The architecture of the baselines with Bayesian Aggregation (BA) follow \citet{volpp2020bayesian} with separate MLPs predicting each neural sufficient statistic (or latent observation and observation variance as elaborated in \citet{volpp2020bayesian}).
The 2nd neural sufficient statistic (i.e. observation noise) is processed using a lower-bounded sigmoid, identical to how the latent variance is processed in the baselines with mean-pooling.
Following \citet{volpp2020bayesian}, a Gaussian prior with fixed parameters is used: $\vmu_0=\vzero, \MSigma_0 = \MI$.
We use MLPs with 64 hidden units and 4 layers.

For our proposed robust Bayesian Aggregation, we extend the aforementioned BA implementation. 
The gamma prior parameters are set as follows: $a_0 = b_0 = 10^{-6} \cdot D$ and $c_0 = 10^{-2} \cdot D$. This is similar to \citet{tipping2005variational} but appropriately scaled by the latent dimensionality. 
For the image completion experiment, the depth of all MLPs is increased by 1.

The models evaluated in \cref{sec:exp_mba} are trained using 10 latent samples and those evaluated in \cref{sec:exp_rba} are trained using 5 latent samples.
We use more latent samples for the mixture experiments as suggested in \citet{wang2022moe}.
Following \citet{lee2020bootstrapping}, all models are optimized using ADAM with initial learning rate $5 \cdot 10^{-4}$ and cosine annealing scheme for the learning rate schedule. 
For the 1-D regression experiment, models are trained for 100,000 steps where each step consists of 16 tasks.
For the image completion experiment, models are trained for 200 epochs with batches of 100 images.

To evaluate the models, we compute the posterior predictive log-likelihood and RMSE using a Monte-Carlo approximation. Following \citet{kim2019attentive} we also evaluate the criterion by using the context points as targets as well. This gives an indication of how well the model is fitting the context points (reconstruction error).

\end{document}

%% file: macros.tex
\newcommand{\mathbold}[1]{\bm{#1}}
\newcommand{\mbf}[1]{\mathbf{#1}}

\newcommand{\T}{\top}    %
\newcommand{\E}{\mathbb{E}}    %
\newcommand{\N}{\mathcal{N}}   %
\newcommand{\G}{\mathcal{G}}   %
\newcommand{\U}{\mathcal{U}}   %
\DeclareMathOperator{\tr}{tr}

\newcommand{\KL}[2]{\mathrm{D}_\mathrm{KL}\left[#1\,\|\,#2\right]}

\newcommand{\vbeta}[0]{\mathbold{\beta}}

\newcommand{\veta}[0]{\mathbold{\eta}}
\newcommand{\vmu}[0]{\mathbold{\mu}}

\newcommand{\vsigma}[0]{\mathbold{\sigma}}

\newcommand{\MSigma}[0]{\mathbold{\Sigma}}

\renewcommand{\mid}[0]{\,|\,}

\newcommand{\vzero}[0]{\mathbold{0}}

\newcommand{\va}{\mbf{a}}
\newcommand{\vb}{\mbf{b}}

\newcommand{\vf}{\mbf{f}}

\newcommand{\vm}{\mbf{m}}

\newcommand{\vr}{\mbf{r}}

\newcommand{\vx}{\mbf{x}}
\newcommand{\vy}{\mbf{y}}
\newcommand{\vz}{\mbf{z}}

\newcommand{\MI}{\mbf{I}}

\newcommand{\MT}{\mbf{T}}

\newcommand{\MV}{\mbf{V}}

\newcommand{\rnd}[1]{\left(#1\right)}
\newcommand{\sqr}[1]{\left[#1\right]}

{}
{}
\newcommand{\data}{\mathcal{D}}

\newcommand{\ang}[1]{\left\langle#1\right\rangle}
\newcommand{\pms}[1]{\ensuremath{{\scriptstyle\pm #1}}}

\definecolor{darkgreen}{rgb}{0.0, 0.5, 0.0}

\usepackage{xspace}
\newcommand{\eg}{\textit{e.g.}\@\xspace}

\newcommand{\etc}{\textit{etc.}\@\xspace}

%% file: figs/tikz/generative.tex
\begin{tikzpicture}
    \node[obs]                                (y_c)   {\Large $y_{c,i}^{(l)}$};
    \node[obs, above=of y_c]                  (x_c)   {\Large $x_{c,i}^{(l)}$};
    \node[obs, right=of y_c, xshift=1.5cm]    (y_t)   {\Large $y_{t,i}^{(l)}$};
    \node[obs, above=of y_t]                  (x_t)   {\Large $x_{t,i}^{(l)}$};
    \node[latent, above=of x_c, xshift=1.2cm,minimum size=1cm] (z)     {\Large $z_l$};
    \factor[above=of z, yshift=0.3cm]         {theta} {\Large $\theta$} {} {};
    \factor[left=of z, xshift=-3.5cm]          {phi_nn} {\Large $\phi$} {} {};

    \edge {x_c} {y_c} ; %
    \edge {x_t} {y_t} ; %
    \edge[] {z} {y_t,y_c} ; %
    \factoredge {} {theta} {z} ; %
    \edge[draw=red,dashed,bend left,line width=0.3mm] {x_c} {z} ;
    \edge[draw=red,dashed,bend right,line width=0.3mm] {y_c} {z} ;
    \edge[draw=red,dashed,line width=0.3mm] {phi_nn} {z} ;
    \plate {cntxt} {(y_c)(x_c)} {\Large $i=1,\ldots,N_{l,c}$} ;
    \plate {trgt} {(y_t)(x_t)} {\Large $i=1,\ldots,N_{l,t}$} ;
    \plate {} {(cntxt)(trgt)(z)} {\Large $l=1,\ldots,L$} ;
\end{tikzpicture}

%% file: tables/mog_emnist.tex
\begin{tabular}{lrrrr || rrrr}
    \toprule
    & \multicolumn{4}{c||}{\textbf{Predictive Log-Likelihood} $\uparrow$} & \multicolumn{4}{c}{\textbf{RMSE} $\downarrow$} \\
    &  \multicolumn{2}{c}{Seen classes (0-9)} & \multicolumn{2}{c||}{Unseen classes (10-46)} &  \multicolumn{2}{c}{Seen classes (0-9)} & \multicolumn{2}{c}{Unseen classes (10-46)}\\
    \cmidrule[0.2pt]{2-5} \cmidrule[0.2pt]{6-9} &  context & target & context & target  &  context & target & context & target \\
    \midrule     
     NP              & $0.701 \pms{0.065}$ & $0.589 \pms{0.061}$ & $0.567 \pms{0.059}$ & $0.405 \pms{0.039}$ & $0.201 \pms{0.018}$ & $0.218 \pms{0.014}$ & $0.244 \pms{0.014}$ & $0.265 \pms{0.009}$ \\
     NP+SA           & $\mathbf{0.977} \pms{0.006}$ & $\mathbf{0.840} \pms{0.005}$ & $0.823 \pms{0.007}$ & $0.609 \pms{0.008}$ & $0.127 \pms{0.002}$ & $\mathbf{0.165} \pms{0.001}$ & $0.177 \pms{0.002}$ & $0.224 \pms{0.002}$ \\
     \hline
     NP-BA           & $0.866 \pms{0.097}$ & $0.708 \pms{0.076}$ & $0.749 \pms{0.118}$ & $0.537 \pms{0.093}$ & $0.154 \pms{0.027}$ & $0.193 \pms{0.014}$ & $0.193 \pms{0.033}$ & $0.238 \pms{0.018}$ \\
     NP-mBA (K=2) & $0.952 \pms{0.121}$ & $0.778 \pms{0.083}$ & $0.855 \pms{0.143}$ & $0.623 \pms{0.097}$ & $0.128 \pms{0.033}$ & $0.181 \pms{0.017}$ & $0.162 \pms{0.038}$ & $0.221 \pms{0.020}$ \\
     NP-mBA (K=3) & $0.953 \pms{0.108}$ & $0.777 \pms{0.079}$ & $0.857 \pms{0.132}$ & $0.623 \pms{0.103}$ & $0.128 \pms{0.031}$ & $0.180 \pms{0.015}$ & $0.162 \pms{0.038}$ & $0.221 \pms{0.020}$ \\
     NP-mBA (K=5) & $0.975 \pms{0.083}$ & $0.792 \pms{0.064}$ & $\mathbf{0.883} \pms{0.105}$ & $\mathbf{0.642} \pms{0.084}$ & $\mathbf{0.122} \pms{0.025}$ & $0.177 \pms{0.011}$ & $\mathbf{0.155} \pms{0.031}$ & $\mathbf{0.217} \pms{0.016}$ \\
    \hline
\end{tabular}

%% file: tables/runtime_image.tex
\begin{tabular}{lccc}
\toprule
& \# Parameters & Training (s) \\
\midrule
NP & 248,962 & 8.1 \\
NP+SA & 282114 & 11.2 \\
\hline
NP-BA & \multirow{3}{*}{166,914} & 10.0 \\
NP-mBA & & 16.3 (*) \\
NP-rBA & & 16.1 \\
\hline\hline
& & \scriptsize{(*) 10 latent samples}
\end{tabular}